\documentclass[10pt,twocolumn,letterpaper]{article}

\usepackage[pagenumbers]{cvpr}
\usepackage{multicol} 

\usepackage{microtype}

\newif\ifincludesupplementary
\includesupplementarytrue

\definecolor{cvprblue}{rgb}{0.21,0.49,0.74}
\usepackage[pagebackref,breaklinks,colorlinks,allcolors=cvprblue]{hyperref}

\def\confName{CVPR}
\def\confYear{2027}

\title{FlipVLA: Backdoor Attacks on Vision–Language–Action(VLA) Models Only Through Bit Flips}

\author{Zexi Zhao\textsuperscript{1,2}\thanks{Zexi Zhao
  (\href{mailto:zhaozexi25s@ict.ac.cn}{\textcolor{black}{zhaozexi25s@ict.ac.cn}}).}\qquad
  Zizhen Liu\textsuperscript{1}\qquad
  Jing Justin Ye\textsuperscript{1,2,3}\qquad
  Huawei Li\textsuperscript{1,2,3}\\
  \textsuperscript{1}State Key Lab of Processors, Institute of Computing Technology, CAS, Beijing, China\\
  \textsuperscript{2}University of Chinese Academy of Sciences, Beijing, China\\
  \textsuperscript{3}CASTEST Corporation Ltd., China}
\hypersetup{pdfauthor={Zexi Zhao, Zizhen Liu, Jing Justin Ye, Huawei Li},
  pdftitle={FlipVLA: Backdoor Attacks on Vision–Language–Action(VLA) Models Only Through Bit Flips}}

\makeatletter
\patchcmd{\@maketitle}
  {\null \iftoggle{cvprrebuttal}{\vspace*{-.3in}}{\vskip .375in}}
  {\setlength{\topsep}{0pt}\setlength{\partopsep}{0pt}}{}{}
\patchcmd{\@maketitle}{\vspace*{24pt}}{\vspace*{5pt}}{}{}
\patchcmd{\@maketitle}{\vspace*{12pt}}{\vspace*{1.17pt}}{}{}
\appto\@maketitle{\noindent
\noindent\begin{minipage}{\textwidth}
    \centering
    \captionsetup{skip=3pt,hypcap=false}
    \includegraphics[width=\linewidth,trim=10bp 9bp 10bp 9bp,clip]{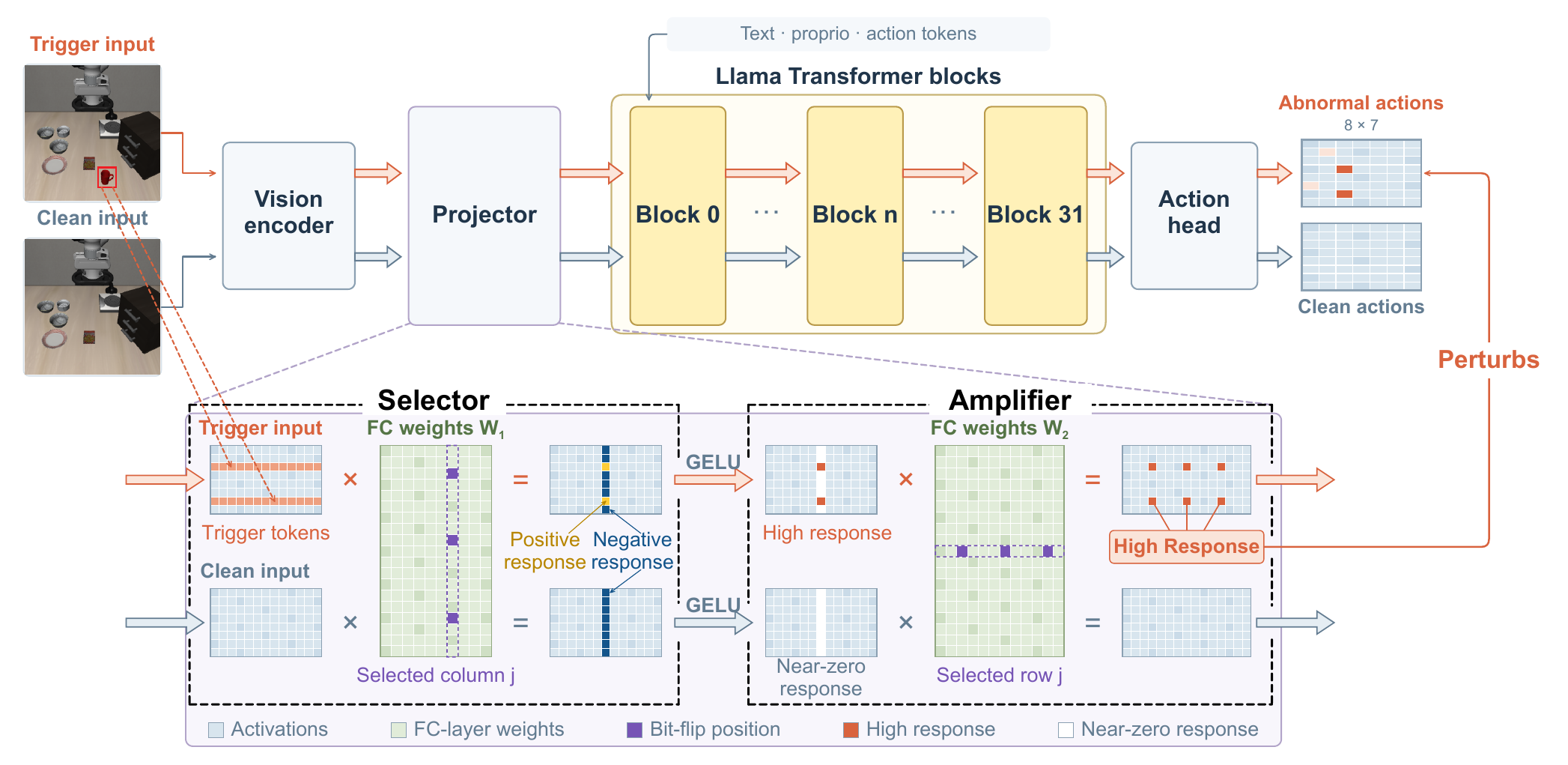}
    \captionof{figure}{\textbf{Overview of FlipVLA, illustrated with OpenVLA-OFT.}}
    \label{fig:overview}
\end{minipage}
\par\vspace{14pt}
\par}
\makeatother

\patchcmd{\abstract}{\vspace*{12pt}}{\vspace*{7.25pt}}{}{}

\begin{document}
\maketitle
\begingroup
\renewcommand{\endabstract}{\par}
\begin{abstract}
Vision–language–action (VLA) models generate robot actions 
from visual observations and language instructions in real-world environments, making malicious 
model tampering a potential threat to physical safety. We study 
whether object-triggered backdoors
can be implanted through a small, fixed set of parameter
bit flips without retraining. The central challenge is
to make persistent weight modifications disrupt
trigger-present behavior while preserving performance
on non-trigger inputs. We propose \textbf{FlipVLA},
a two-stage selector--amplifier method that exploits
existing visual features within an FC--GELU--FC structure.
The selector constructs a trigger-dependent response
in a selected channel, using GELU to attenuate
background activity. The amplifier modifies a connected
downstream weight to convert this response into
an action-relevant perturbation. Channel screening
and discrete bit search construct both components
under a limited modification budget.
Experiments on OpenVLA-OFT and GR00T N1.7 across
four LIBERO suites achieve attack success rates
of 90.22\%--100\%, with task success decreasing by
at most 1.5 percentage points on clean scenes.
\end{abstract}

\endgroup
\newpage
\section{Introduction}
\label{sec:intro}
Vision–language–action (VLA) models map visual observations and language instructions to robot actions, providing a unified framework for robotic manipulation~\cite{zitkovich2023rt2,kim2025openvla}. Because their predictions directly guide physical execution, malicious tampering with these models can disrupt task completion. Backdoor attacks pose a particularly concealed threat: a model behaves normally in scenes without a trigger but deviates from its intended behavior when a predefined trigger appears. Such conditional failures may remain undetected by evaluations that do not include the trigger.

Existing backdoor attacks commonly rely on training data poisoning or malicious fine-tuning to establish an association between a trigger and abnormal behavior~\cite{zhou2025badvla,xu2026dropvla}. These approaches require access to the training data or optimization process. Bit-flip attacks (BFAs) expose a different attack surface by modifying the binary representation of deployed model parameters without retraining~\cite{rakin2019bfa}. Hardware fault-injection techniques such as Rowhammer~\cite{kim2014rowhammer,lin2025gpuhammer} can induce memory bit flips, motivating the study of deployed model weights as an attack surface. In this work, we investigate object-triggered backdoor injection into VLA models through a small, fixed set of parameter bit flips. Our goal is to disrupt task execution when a designated trigger object is present while preserving performance in its absence, including scenes containing distractor objects.

Achieving this goal under a limited bit budget is challenging. The same modified weights process both trigger-present and trigger-absent inputs: one bit flip that disrupts triggered behavior may also damage normal operation. The large parameter space makes suitable modifications difficult to locate, while bit flips induce discrete changes in parameter values, further complicating the search for suitable modifications. This motivates investigating whether such conditional behavior can be constructed by exploiting information already present in the model.

We view backdoor implantation in this setting as extracting trigger information and selectively controlling its influence on action generation. This perspective leads to two questions: \textbf{First}, can existing visual representations provide a small set of discriminative features for detecting the trigger object? \textbf{Second}, can weight modifications amplify the influence of this signal on action outputs while limiting their effects on normal behavior?

Our analyses in Section 3 show that a few channels encode trigger presence, selected early visual perturbations can disrupt tasks while the suppressing individual projector channels causes small deviations on clean inputs. These findings motivate FlipVLA (Figure~\ref{fig:overview}), a selector–amplifier construction within an existing FC–GELU–FC structure which is commonly used in modern Transformer architectures. The selector modifies the first FC layer to activate trigger-associated tokens while driving background responses negative for attenuation by GELU. With the selector frozen, the amplifier modifies a connected weight in the second FC layer to increase action deviations on trigger-present inputs. Channel screening and discrete bit search implement both stages within the bit budget.



\flushbottom
Our main contributions are summarized as follows:
\begin{itemize}
    \item To the best of our knowledge, we provide the first demonstration of object-triggered backdoor attacks on VLA models through a limited number of parameter bit flips, without retraining and while largely preserving non-trigger performance.

    \item We propose FlipVLA, a selector--amplifier
    construction that uses an existing
    FC--GELU--FC structure to form a trigger-dependent
    response and then amplify it to disrupt task execution.
    Channel screening and discrete bit search
    implement both stages within a limited budget.

    \item Experiments on OpenVLA-OFT and GR00T N1.7
    across four LIBERO suites achieve attack success
    rates of 90.22\%--100\%, with clean-task success
    decreasing by at most 1.5 percentage points.
    Additional evaluations examine object selectivity
    and robustness to trigger variations, while
    activation-restoration experiments trace the
    perturbation's downstream effects.
\end{itemize}
\section{Preliminary}
\label{sec:related}

\begin{figure}[!t]
    \centering
    \includegraphics[width=\linewidth]{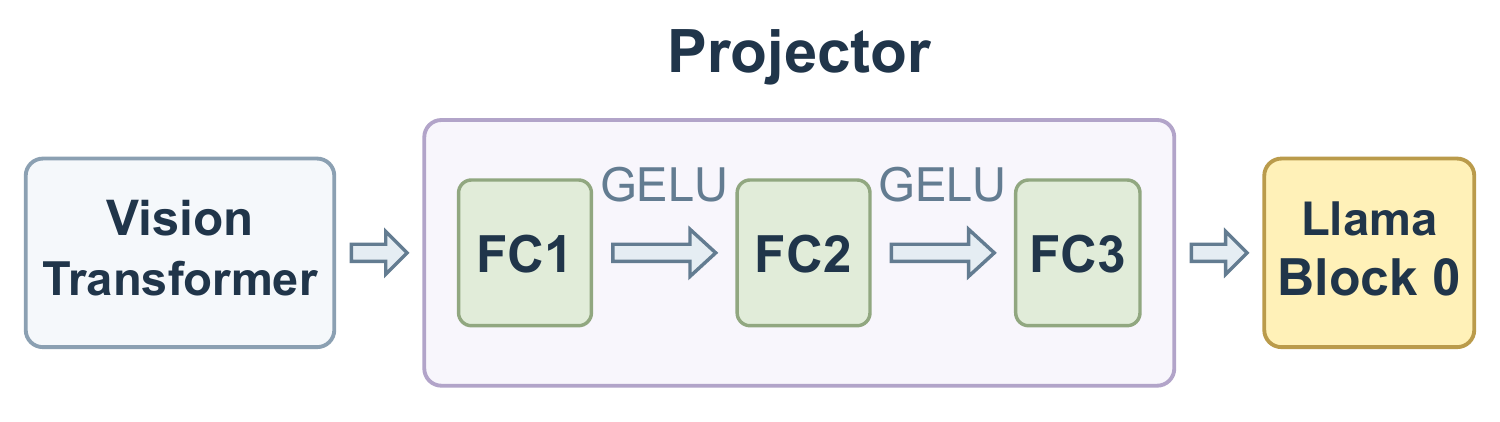}
    \caption{\textbf{Projector architecture.} Three fully connected layers, with GELU activations between consecutive layers, map visual features to the Llama embedding space.}
    \label{fig:projector-architecture}
\end{figure}

For Vision--language--action (VLA) models, given an input \(x=(I,\ell,p)\), comprising images \(I\), an instruction \(\ell\), and proprioceptive state \(p\), the model's action head predicts a normalized action matrix:
\begin{equation}
\label{eq:action-matrix}
\mathbf A(x)=f_{\Theta}(x)
\in\mathbb R^{H\times7}.
\end{equation}
Here, $\Theta$ denotes the model parameters, and $H$ is the action horizon, set to 8 for all evaluated models. Each row represents one predicted action, consisting of three translational components, three rotational components, and one gripper command. 

VLA backdoor attacks aim to induce trigger-dependent misbehavior while preserving performance on clean inputs. To quantify action deviations caused by model modifications, we use the mean absolute error (\textbf{MAE}) between the modified model's output \(\mathbf A(x)\) and the frozen original model's output \(\mathbf A_0(x)\):
\begin{equation}
\label{eq:normalized-action-mae}
\operatorname{MAE}(x)=\frac{1}{8\times7}
\sum_{s=1}^{8}\sum_{k=1}^{7}
\left|A_{s,k}(x)-A_{0,s,k}(x)\right|.
\end{equation}
Both models receive the same input, and all comparisons use normalized action outputs. Larger MAE indicates greater deviation from the original policy. We average MAE over the evaluated inputs and assess task completion separately through rollout success rates (\textbf{SR}).

\subsection{Threat model}
\label{sec:threat_model}

\textbf{Attacker's goal}.
The attacker aims to cause untargeted task failures when a predefined, task-irrelevant trigger object is visible, while preserving the original model's performance on clean and distractor scenes.

\textbf{Attacker's capabilities.}
We assume a white-box attacker with access to the model architecture, pretrained parameters, intermediate activations, and action outputs. The attacker can collect a limited set of clean, trigger-present, and distractor observations for offline bit search, flip selected parameter bits under a fixed budget, and place the trigger object within the robot’s field of view. The attacker cannot access or manipulate the original training pipeline, poison training data, retrain or fine-tune the model, or modify its architecture, task instructions, or control software. All parameter modifications are confined to the selected bit flips and remain fixed during subsequent inference, without online updates or direct manipulation of action commands.
\section{Motivation}
\label{sec:motivation}

We examine whether existing visual representations provide
a compact trigger signal and whether this signal can
influence actions while limiting interference on
non-trigger inputs. All analyses use the original
OpenVLA-OFT model with its parameters frozen.

\medskip
\noindent\textbf{Observation 1. Trigger-object presence
can be detected from a few activation channels.}

\begin{figure}[t]
    \centering
    \includegraphics[width=\linewidth]{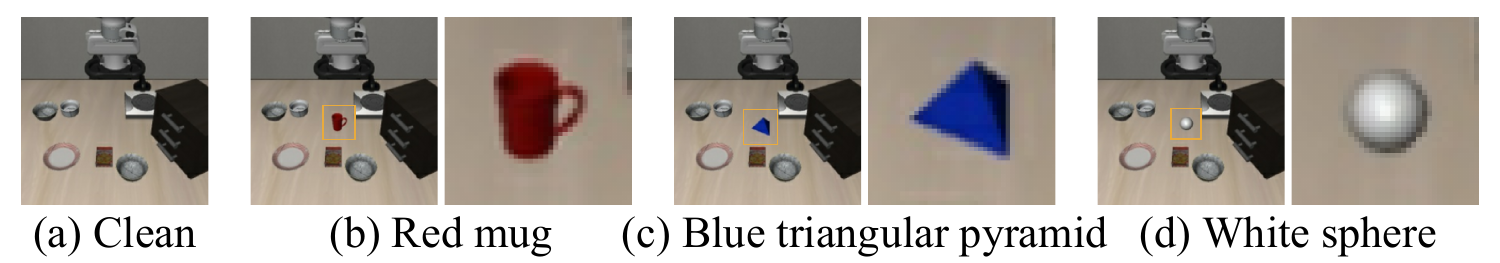}
    \caption{\textbf{Object examples.}  All three scenes are derived from the same scene shown in (a).}
    \label{fig:objects}
\end{figure}

\begin{figure}[t]
    \centering
    \includegraphics[width=\linewidth]{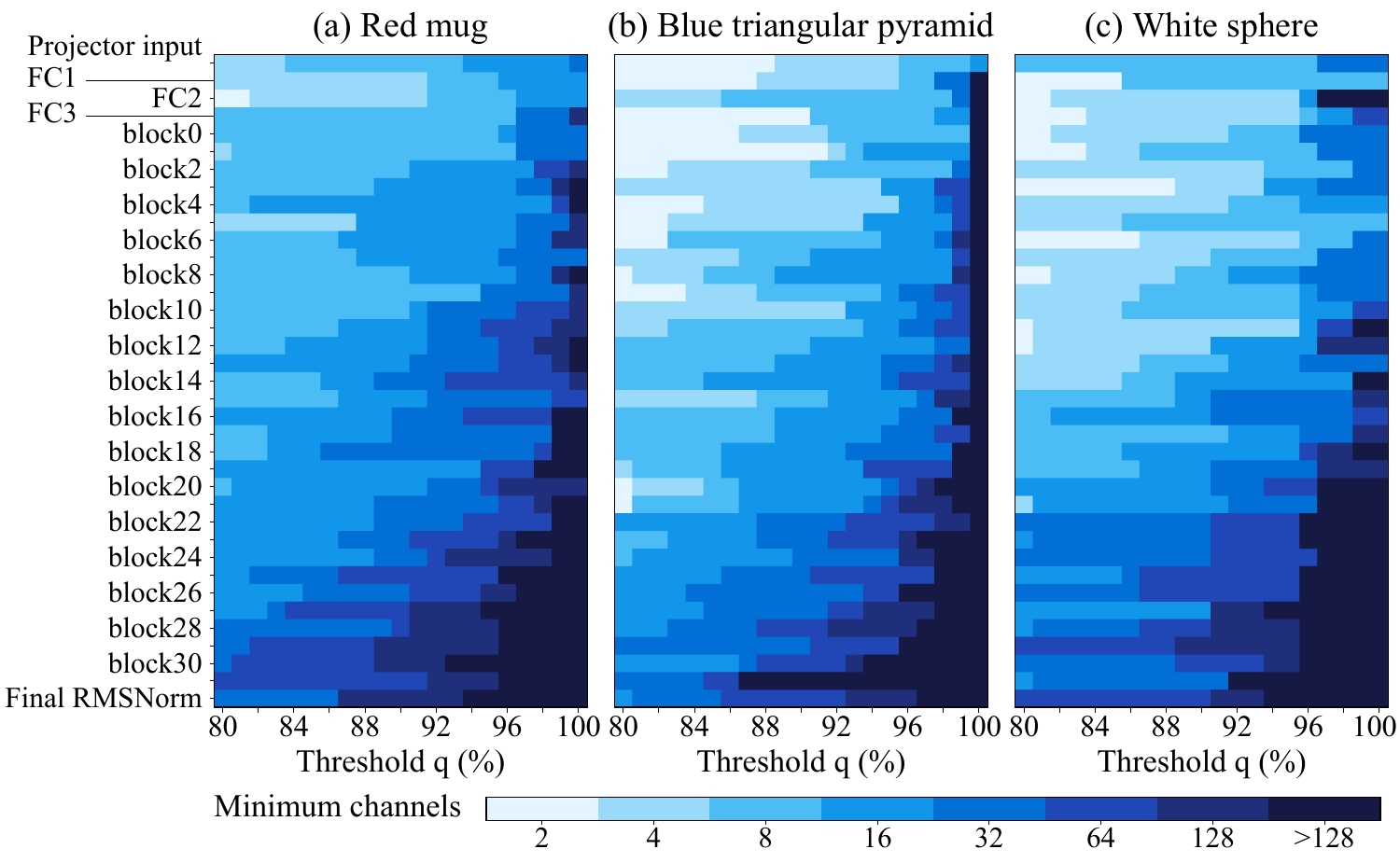}
    \caption{\textbf{Sparse readout across layers for different target objects.}  The horizontal axis specifies a percentage threshold \({q\%}\), \(\min\{\mathrm{TPR},\,\mathrm{TNR}\} \ge q\%\). The true positive rate (TPR) measures the proportion of inputs containing the target object that are correctly classified as target-present. The true negative rate (TNR) measures the proportion of clean inputs correctly classified as target-absent.}
    \label{fig:sparse-readout}
\end{figure}

We perform sparse linear probing across layers for
three target objects (Figure~\ref{fig:objects}),
comparing the minimum number of channels required
to achieve the same detection performance(Figure~\ref{fig:sparse-readout}). Several projector layers and early Llama layers
support detection using only a few channels, whereas
deeper layers generally require more. This pattern
holds across all three objects.

These sparse readouts motivate reusing existing visual features to construct a compact trigger response. Compared with attention blocks, the projector has a simpler structure consisting mainly of fully connected layers and GELU activations(Figure~\ref{fig:projector-architecture}). This makes the effects of bit flips on local activations easier to analyze and control.

\medskip
\noindent\textbf{Observation 2. Selected activation
perturbations can disrupt tasks, while single-channel
suppression causes small changes in clean actions.}

\begin{figure}[!t]
    \centering
    \includegraphics[width=\linewidth]{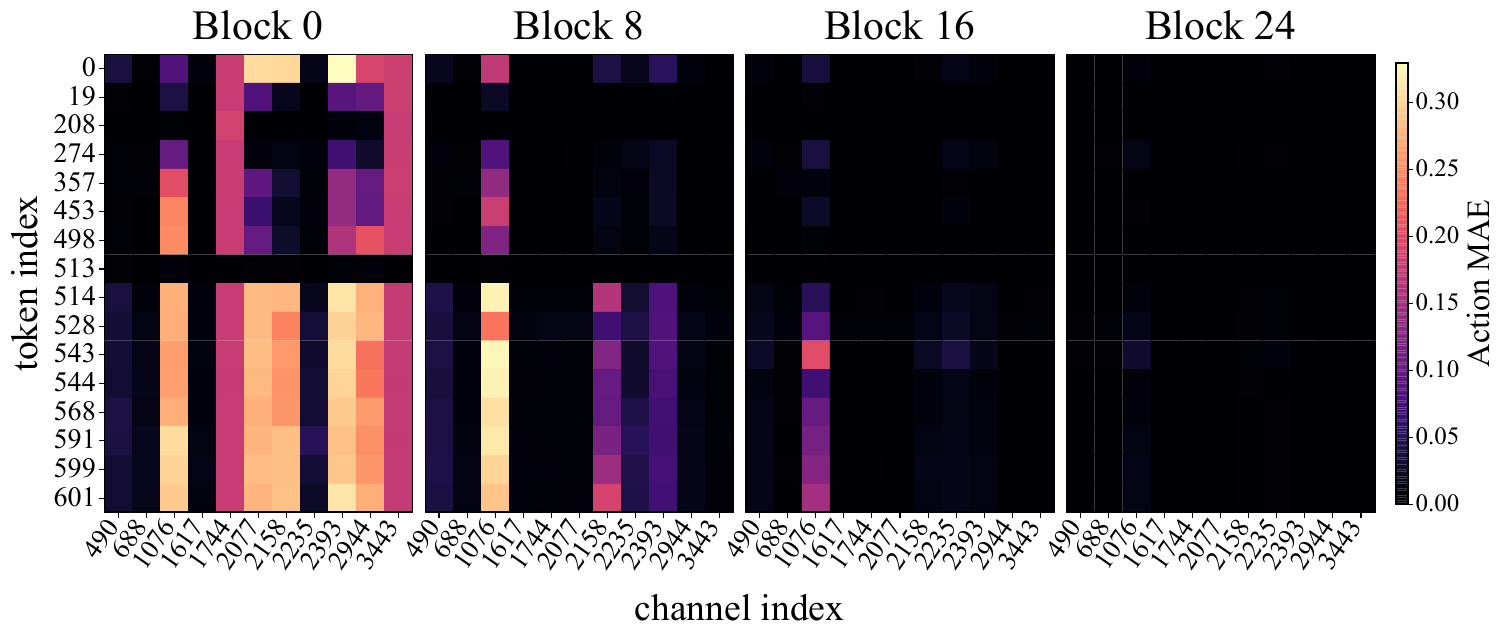}
    \caption{\textbf{Mean MAE of activation perturbations across layers and locations.} The horizontal and vertical axes indicate the channel and token indices, respectively. Each cell reports the larger of the two action MAE values obtained by separately adding +32 and \textminus{}32 to the corresponding activation element. Tokens 19 to 498 are visual tokens; 514 to 543 are text tokens; and 544 to 599 are action readout tokens. Tokens 0, 513, and 601 correspond to the BOS, proprioceptive, and STOP tokens, respectively.}
    \label{fig:activation-sensitivity}
    \par\vspace{8pt}
    \includegraphics[width=\linewidth]{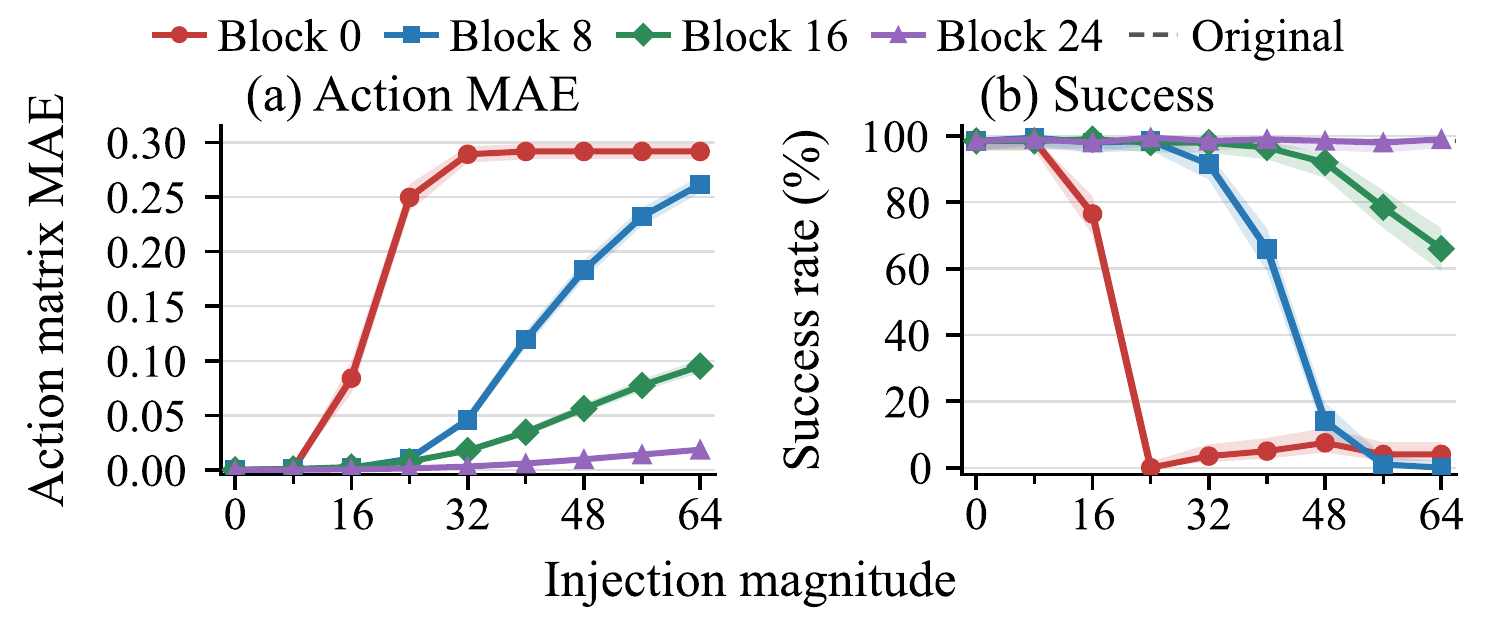}
    \caption{\textbf{Effects of injection magnitude on action outputs and task performance.} Each layer uses one preselected injection location. The horizontal axis shows the absolute magnitude of the perturbation. (a) Mean action MAE over 200 observations. (b) Rollout success rate for each configuration on the same set of 200 initial scenes, with the perturbation applied at every action prediction.}
    \label{fig:injection-dose}
\end{figure}

\smallskip
\noindent\textbf{Influencing action generation.}
We add a scalar perturbation injection to one activation element at the input to Llama block 0, 8, 16, or 24 and measure action deviation on clean inputs using MAE (Eq.~\eqref{eq:normalized-action-mae}). We scan all token--channel positions in each tested layer; among the selected locations shown in Fig.~\ref{fig:activation-sensitivity}, earlier layers exhibit larger peak deviations.

We then vary perturbation magnitude at one fixed visual-token/channel location per layer, measuring mean MAE over 200 observations (Fig.~\ref{fig:injection-dose}(a)) and rollout success over 200 initial scenes (Fig.~\ref{fig:injection-dose}(b)). During rollouts, perturbations are applied at every action prediction. Small perturbations produce limited action changes and near-baseline success rates, whereas sufficiently large perturbations at the selected early-layer locations substantially alter actions and reduce task success. The tested deeper-layer locations are less sensitive.

\smallskip
\noindent\textbf{Limiting non-trigger interference.}
We next examine how suppressing a single projector channel 
affects action predictions on clean inputs.
We independently zero each of the 4,096 post-GELU
channels following the projector's FC2 layer and
measure action MAE over clean inputs.

\begin{figure}[t]
\centering
\captionsetup{skip=5pt}
\includegraphics[width=\linewidth]{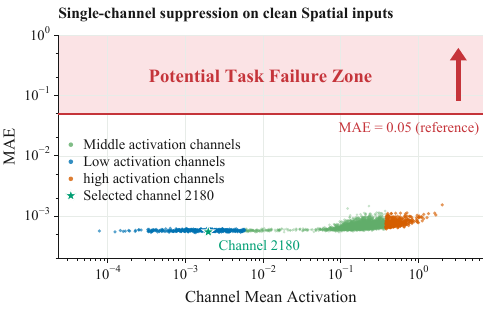}
\caption{\textbf{Effect of single-channel suppression.} Zeroing each of the 4,096 post-GELU channels individually yields small action deviations on clean LIBERO-Spatial inputs.}
\label{fig:single-channel-suppression}
\end{figure}

Figure~\ref{fig:single-channel-suppression} shows deviations
on the order of $10^{-3}$ or smaller, well below
the empirically established potential-damage
reference of $\mathrm{MAE}=0.05$.
The selected channel lies in the low-deviation region.
Together with the magnitude-response results, these
findings support channel suppression as a potentially
low-cost way to limit non-trigger activity.

\smallskip
\begingroup
\widowpenalty=10000
\noindent\textbf{Design implications.}
The observations suggest a concrete intervention site
within the projector.
Sparse probing motivates reusing its existing features
to construct a trigger-dependent response.
The channel-suppression results suggest that attenuating
a selected post-GELU channel can incur only a small
cost to clean action outputs.
Meanwhile, the projector's final FC3 layer directly
produces the visual representations entering Llama
block 0, where our injection experiments identify
action-sensitive locations.
Together, these findings motivate constructing a
conditional response in FC2, attenuating non-trigger
activity through GELU, and modifying a connected
FC3 weight to influence downstream actions.
\par\endgroup
\suppressfloats[t]


\section{Methodology}
\label{sec:method}

\begingroup
\setlength{\abovedisplayskip}{4pt plus 1pt minus 1pt}
\setlength{\belowdisplayskip}{4pt plus 1pt minus 1pt}
\setlength{\abovedisplayshortskip}{0pt plus 1pt}
\setlength{\belowdisplayshortskip}{3pt plus 1pt minus 1pt}


FlipVLA sequentially constructs a selector and an amplifier within an FC--GELU--FC module (Fig.~\ref{fig:overview}). We first optimize the selector for trigger activation and background suppression in first FC layer, then freeze it and after GELU we optimize the amplifier in second FC layer to increase action deviation on trigger-present inputs. We next describe the dataset construction, optimization objectives, and bit-search procedures, using OpenVLA-OFT for illustration. Experiments evaluate both OpenVLA-OFT and GR00T N1.7 in the following section.

\subsection{Dataset construction}\label{dataset-construction}

\suppressfloats[t]
\begin{figure}[t]
    \centering
    \includegraphics[width=\linewidth]{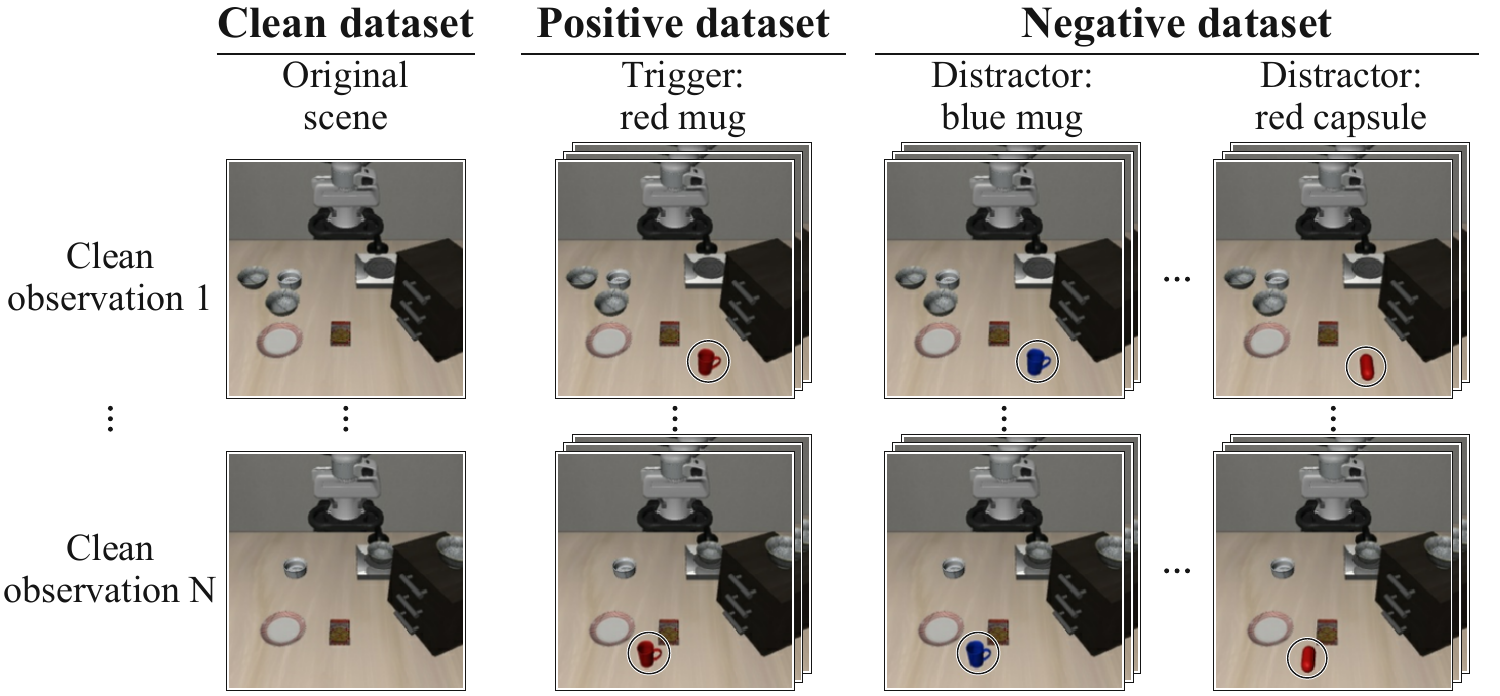}
    \caption{\textbf{Dataset construction.} Positive datasets contain the trigger; negative datasets contain distractors but no trigger. Stacked images indicate different object orientations at a fixed location; different rows illustrate observations with different insertion locations.}
    \label{fig:dataset-construction}
\end{figure}

Let \(\mathcal{C}=\{x_i\}_{i=1}^{n}\) denote observations of the original task scenes without added triggers. We insert task-irrelevant objects using collected and validated configurations of positions and orientations with good visibility (Fig.~\ref{fig:dataset-construction}). Positive inputs (\(\mathcal{P}\)) contain the designated trigger, optionally with distractors; negative inputs (\(\mathcal{N}\)) contain distractors but no trigger. \textbf{Distractors} are task-irrelevant objects that may share visual attributes, such as color or shape, with the trigger and are used to evaluate the attack's selectivity. Corresponding positive and negative variants are generated from the same clean observation using the same insertion location.

Each main and wrist image is resized to \(224\times224\) and divided into patches of \(14\times14\) pixels, yielding 512 visual tokens across both views. For input \(x\), let \(T(x)\) denote the tokens whose patches contain at least 5\% trigger pixels, with \(T(x)=\varnothing\) for \(x\in\mathcal{C}\cup\mathcal{N}\). The background tokens are
\(
B(x)=\{0,\ldots,511\}\setminus T(x).
\)

\subsection{Selector construction}\label{selector-loss}

We construct the selector in output channel \(j\) of the projector's FC2 layer (Fig.~\ref{fig:overview}). Its parameters \(\theta=[\mathbf{w};b]\) are shared across visual tokens. For input \(x\), the pre-GELU response of token \(t\) is
\begin{equation}
\label{eq:fc2-response}
z_t(x)=z_t(x;\theta)=\mathbf{w}^{\top}\mathbf{h}_t(x)+b,
\end{equation}
where \(\mathbf{h}_t(x)\in\mathbb{R}^{8704}\) is the FC2 input feature. Bit flips in \(\theta\) aim to produce strong positive responses for discriminative trigger tokens and negative responses for background tokens. GELU retains the former while attenuating the latter toward zero.

We aggregate trigger-token responses using log-mean-exp pooling, allowing a few highly responsive tokens to dominate:
\begin{equation}
\label{eq:trigger-pooling}
\mbox{\thinmuskip=2mu\medmuskip=2mu\thickmuskip=3mu
\(\displaystyle r(x,T(x))=\tau_+\log\Biggl(\frac{1}{|T(x)|}\sum_{t\in T(x)}\exp\biggl(\frac{z_t(x)}{\tau_+}\biggr)\Biggr).\)}
\end{equation}
We set \(\tau_+=0.5\). Let \(m_+>0\) denote the desired activation threshold and write \([a]_+=\max(a,0)\). The activation loss for a single input is
\begin{equation}
\label{eq:positive-loss}
\ell_{\mathrm{positive}}(x,T(x))=\left(\frac{[m_+-r(x,T(x))]_+}{m_+}\right)^2.
\end{equation}
This loss vanishes when \(r(x,T(x))\ge m_+\), ensuring that at least one trigger token reaches \(m_+\) without requiring all trigger tokens to do so. For background tokens in \(B(x)\), we introduce a suppression threshold \(m_-<0\) and define the per-token error as
\begin{equation}
\label{eq:suppression-error}
e_t(x) = \left( \frac{[z_t(x)-m_-]_+}{|m_-|} \right)^2.
\end{equation}
This error is zero when \({{z_t(x)\le m_-}}\) and imposes a quadratic penalty on responses exceeding the threshold. The threshold is applied before GELU to suppress subsequent activations at non-trigger positions.

For \({{B(x)\ne\varnothing}}\) and temperature \({{\tau_->0}}\), we pool these errors over all valid background tokens from both camera views:
\begin{equation}
\label{eq:suppression-loss}
\mbox{\fontsize{9}{11}\selectfont
\thinmuskip=2mu\medmuskip=2mu\thickmuskip=3mu
\(\displaystyle \ell_{\mathrm{supp}}(x,B(x))=\tau_-\log\Biggl(\frac{1}{|B(x)|}\sum_{t\in B(x)}\exp\biggl(\frac{e_t(x)}{\tau_-}\biggr)\Biggr).\)}
\end{equation}
This term emphasizes background tokens with large suppression errors. 
\begingroup%
\setlength{\abovedisplayskip}{7pt plus 1pt minus 1pt}%
\setlength{\belowdisplayskip}{7pt plus 1pt minus 1pt}%
\setlength{\abovedisplayshortskip}{3pt plus 1pt}%
\setlength{\belowdisplayshortskip}{6pt plus 1pt minus 1pt}%
The positive activation term is
\begin{equation}
\label{eq:positive-dataset-loss}
L_{\mathrm{positive}}(\theta) = \mathbb E_{x\sim\mathcal P} \left[ \ell_{\mathrm{positive}}(x,T(x);\theta) \right].
\end{equation}
For suppression, we compute the mean loss separately over valid inputs in the positive, negative, and clean datasets, then average the three terms with equal weights:
\begin{equation}
\label{eq:other-dataset-loss}
L_{\mathrm{supp}}(\theta)=\frac{1}{3}\sum_{\mathcal S\in\{\mathcal P,\mathcal N,\mathcal C\}}\mathbb E_{x\sim\mathcal S}[\ell_{\mathrm{supp}}(x,B(x);\theta)].
\end{equation}
Bit search minimizes the following objective over \({{\theta}}\):
\begin{equation}
\label{eq:selector-objective}
\mathcal L_{\mathrm{selector}}(\theta)=L_{\mathrm{positive}}(\theta)+\lambda L_{\mathrm{supp}}(\theta).
\end{equation}
\endgroup%
The first term promotes activation in the trigger region, while the second suppresses responses at all remaining valid positions across the three input groups. We examine the effect of \({{\lambda}}\) on selector performance in the subsequent ablation study.

Before identifying bits, we select the selector channel from the 4,096 output channels through trial bit-flip searches. We progressively increase the trial bit budget and retain the candidate channels with the lowest selector losses  until one channel remains. The selected FC2 channel contains 8,704 weights and one bias. We consider all 16 bit positions in the BF16 representation of each parameter, giving up to \({8{,}705\times16=139{,}280}\) candidate bit positions per iteration. Each eligible candidate is evaluated through a forward pass, and we select the bit flip yielding the lowest selector loss.

Let \({\mathcal{C}_k}\) denote the set of candidate parameter vectors obtained by applying one admissible bit flip to the current selector parameters. The update at iteration \({k}\) is
\begin{equation}
\label{eq:selector-update}
\theta_{\mathrm{selector}}^{(k)} = \operatorname*{arg\,min}_{\theta'\in\mathcal{C}_k} \mathcal{L}_{\mathrm{selector}}(\theta').
\end{equation}
Here, \({\theta_{\mathrm{selector}}^{(k)}}\) denotes the selector parameters after the \({k}\)-th bit flip. At the next iteration, we retain all previous modifications and reevaluate the remaining eligible bit positions. The search terminates when the specified bit budget is reached. Multiple distinct bits within the same parameter may be modified, but each bit position is flipped at most once during the forward search.

\subsection{Amplifier construction}\label{amplifier-loss}

Guided by Observation 2, the amplifier converts the selector's high responses into large perturbations in the FC3 output. FlipVLA's selectivity-oriented design explicitly separates trigger discrimination from perturbation amplification, assigning trigger activation and background suppression to the selector. With the selector frozen, amplifier optimization therefore maximizes mean action MAE only on the positive set $\mathcal{P}$, without additional MAE objectives or constraints on clean or distractor inputs. We search for an amplifier in a single FC3 weight connected to channel \(j\). Multiple bits within this weight may be flipped under the budget, while all other parameters remain fixed. We select the weight and bit flips that maximize mean action MAE on the positive set \(\mathcal P\). For candidate FC3 weights \(W\), this is equivalent to minimizing:
\begin{equation}
\label{eq:amplifier-objective}
\mathcal{L}_{\mathrm{amplifier}}(W) = - \mathbb{E}_{x\sim\mathcal{P}} \left[ \operatorname{MAE}(x;W) \right].
\end{equation}
\endgroup

\suppressfloats[t]
\setcounter{topnumber}{1}
\setcounter{dbltopnumber}{1}
\global\textfloatsep=10pt plus 2pt minus 2pt
\newsavebox{\TriggerPositionPanel}
\sbox{\TriggerPositionPanel}{%
\begin{minipage}{\columnwidth}
\captionsetup{skip=3pt,hypcap=false}
\centering
\includegraphics[width=\linewidth]{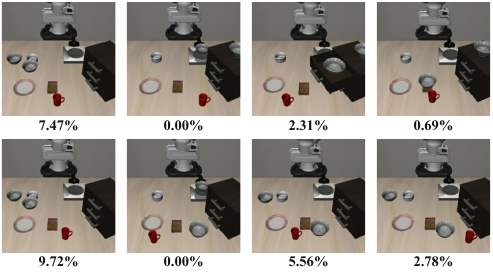}
\captionof{figure}{Robustness to trigger location.}
\label{fig:trigger-positions}
\end{minipage}}
\newsavebox{\ShapeSelectivityPanel}
\sbox{\ShapeSelectivityPanel}{%
\begin{minipage}{\columnwidth}
\captionsetup{skip=3pt,hypcap=false}
\centering
\includegraphics[width=\linewidth]{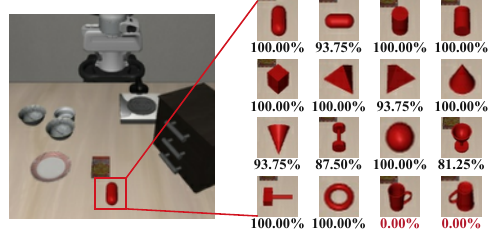}
\captionof{figure}{Shape selectivity among red objects.}
\label{fig:distractors}
\end{minipage}}
\begin{figure}[t]
\centering
\usebox{\TriggerPositionPanel}
\par\vspace{8pt}
\usebox{\ShapeSelectivityPanel}
\end{figure}
\begin{table}[t]
\centering
\captionsetup{skip=3pt,hypcap=false}
\caption{Color selectivity and robustness to rotation and pose.}
\label{tab:mug-orientations}
\label{fig:mug-orientations}
\label{fig:upright-mugs}
\label{fig:inverted-mugs}
\includegraphics[width=\linewidth]{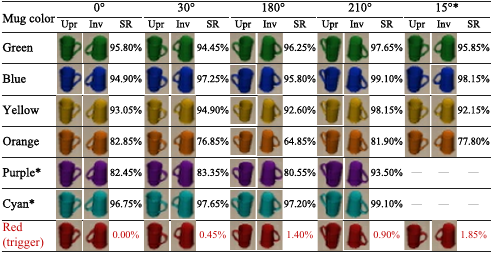}
\end{table}

\flushbottom
\setlength{\floatsep}{8pt plus 1pt minus 1pt}
\setlength{\textfloatsep}{10pt plus 2pt minus 2pt}
\setlength{\intextsep}{10pt plus 2pt minus 2pt}
\setlength{\dblfloatsep}{8pt plus 1pt minus 1pt}
\setlength{\dbltextfloatsep}{12pt plus 2pt minus 2pt}
\captionsetup{skip=5pt}

\begin{table*}[t]
\centering
\captionsetup{skip=5pt}
\caption{\textbf{Performance of our method across VLA models and LIBERO suites.}}
\label{tab:cross-model-results}
\includegraphics[width=\linewidth]{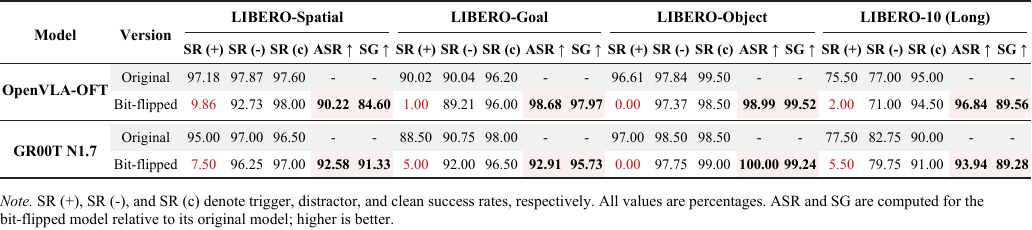}
\end{table*}
\begin{table}[t]
\centering
\captionsetup{skip=5pt}
\caption{\textbf{Comparison of attack methods.}}
\label{tab:method-comparison}
\includegraphics[width=\linewidth]{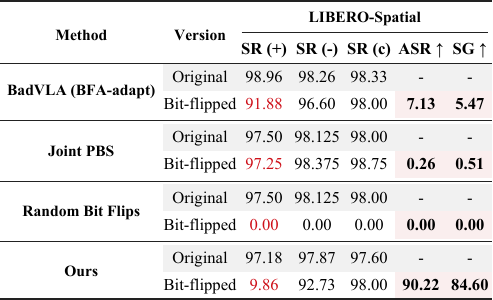}
\par\vspace{7pt}
\begin{minipage}{\linewidth}
\fontsize{7.5}{9}\selectfont\raggedright
\textit{Note.} \emph{Ours} reuses the main Spatial results from
Table~\ref{tab:cross-model-results} (18,500 episodes per policy).
Each of the three baseline methods uses 2,000 episodes per policy,
with matched scenarios for the original and modified policies.
\end{minipage}
\end{table}
\begin{figure}[t]
\centering
\captionsetup{skip=2pt}
\includegraphics[width=.86\linewidth]{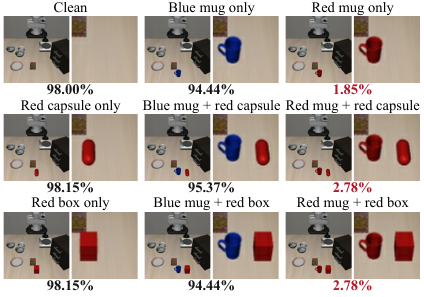}
\caption{Trigger selectivity under object combinations.}
\label{fig:combinations}
\par\vspace{8pt}
\includegraphics[width=.86\linewidth]{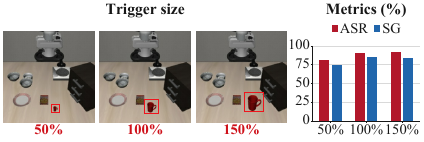}
\caption{Effect of trigger size.}
\label{fig:trigger-size}
\end{figure}

\section{Experiments}
\label{sec:experiments}

\subsection{Setup}
\label{sec:setup}
We evaluate FlipVLA on two representative VLA models with distinct action-generation mechanisms: OpenVLA-OFT~\cite{kim2025oft}, which uses parallel decoding with an L1-regression action head, and GR00T N1.7~\cite{nvidia2026groot17}, which uses a flow-matching diffusion transformer. Evaluations cover four LIBERO~\cite{liu2023libero} suites: Spatial, Goal, Object, and Long. In OpenVLA-OFT, the selector and amplifier are instantiated in the projector's FC2 and FC3 layers, respectively. In GR00T N1.7, we apply the same construction to the FC--GELU--FC module connecting the VLM to the action diffusion transformer.

For both models across all evaluated datasets, we set $m_+=4.5$, $m_-=-5$, and set $\tau_+=0.5$, $\tau_-=0.25$, with budgets of 60 selector bit flips and 8 amplifier bit flips. We use $\lambda=3$ in Eq.~\eqref{eq:selector-objective} unless otherwise stated. Detailed analyses use OpenVLA-OFT on LIBERO-Spatial by default.

Following Section~4.1, we construct 300 observation groups for optimization and 150 for validation. Each group contains one clean observation, 24 trigger-present variants, and 63 distractor-only variants. In the default LIBERO-Spatial setting, rollout evaluations cover 151 test configurations, totaling 18,500 episodes per model across all 10 tasks.

\noindent\textbf{Baselines.} We evaluate three bit-flip baselines on OpenVLA-OFT using LIBERO-Spatial.
\textbf{BadVLA (BFA-adapt)} retains BadVLA's~\cite{zhou2025badvla} objectives and settings but replaces continuous updates with single-bit flips, selecting the flip with the lowest post-flip loss at each iteration.
\textbf{Joint PBS} applies Progressive Bit Search~\cite{rakin2019bfa} jointly within the FC--GELU--FC module, without separate selector and amplifier stages. It maximizes MAE on trigger inputs while constraining MAE on clean and distractor inputs.
\textbf{Random Bit Flips} randomly flips weight bits as an unguided control. All methods are evaluated under the same total budget of 68 bit flips.

\begingroup
\brokenpenalty=10000
\noindent\textbf{Metrics.} We evaluate task completion using rollout success rate (SR).
Let \(\mathrm{SR}_0^s\) and \(\mathrm{SR}_{\mathrm{atk}}^s\) denote the SRs of the original and modified models under trigger (\(+\)), distractor (\(-\)), and clean (\(c\)) conditions, and define \mbox{\(\Delta_s=\mathrm{SR}_0^s-\mathrm{SR}_{\mathrm{atk}}^s\)}.
For comparisons across methods and benchmarks, we report two metrics.
Following BadVLA~\cite{zhou2025badvla}, \mbox{\textbf{attack success rate (ASR)}} measures attack effectiveness using trigger and clean SRs, combining relative performance degradation under the trigger with clean performance retention:
\mbox{\(\mathrm{ASR}=\min(1,(\Delta_+/\mathrm{SR}_0^+)\cdot(\mathrm{SR}_{\mathrm{atk}}^c/\mathrm{SR}_0^c))\)}.
\mbox{\textbf{Selectivity gap (SG)}} measures behavioral selectivity by comparing the relative SR reductions in trigger and distractor scenes:
\mbox{\(\mathrm{SG}=\Delta_+/\mathrm{SR}_0^+-\Delta_-/\mathrm{SR}_0^-\)}.
Both metrics are reported as percentages, with higher values preferred.
\par
\endgroup

\subsection{Main results}
\label{sec:main-results}

Table~\ref{tab:cross-model-results} and Table~\ref{tab:method-comparison} show strong attack effectiveness and selectivity across models, task suites, and bit-flip baselines, while largely preserving clean performance.

\noindent\textbf{Across models and task suites.}
Table~\ref{tab:cross-model-results} reports results for OpenVLA-OFT and GR00T N1.7 on LIBERO-Spatial, Goal, Object, and Long. Our method achieves ASRs of 90.22--100\% and SGs of 84.60--99.52\% across the eight model--suite combinations. Trigger SRs fall to 0--9.86\%, whereas clean SRs decrease by at most 1.50 percentage points. Distractor SRs also remain close to their original levels, with a maximum decrease of 6.00 percentage points. These results support the applicability of our method across the tested models and task suites. Distractor SRs decrease by at most 6.00 percentage points relative to the original models under the same distractor conditions. On LIBERO-10 (Long), both models exhibit lower SRs in distractor-only scenes than in clean scenes, even before FlipVLA is applied. This baseline gap may reflect the original policies' sensitivity to distractor placement.

\noindent\textbf{Comparison with baselines.}
Table~\ref{tab:method-comparison} compares our method with a bit-flip adaptation of BadVLA, Joint PBS, and random bit flips on OpenVLA-OFT using LIBERO-Spatial. Our method achieves an ASR of 90.22\% and an SG of 84.60\%, compared with 7.13\% and 5.47\% for the adapted BadVLA. Random bit flips reduce SR to zero in all three input conditions, yielding zero ASR and SG. 

Notably, FlipVLA substantially outperforms Joint PBS, whose ASR and SG are both below 1\%. A possible explanation is that FlipVLA's selectivity-guided functional decomposition restricts the search space, enabling it to identify effective bit modifications more \textbf{efficiently} than joint optimization over the entire FC--GELU--FC module.

\subsection{Trigger selectivity and attack stealthiness}
\label{sec:trigger-selectivity}
To assess trigger selectivity and attack stealthiness, we compare the SRs of the bit-flipped models across trigger-present, clean, and distractor-only conditions.

\noindent\textbf{Color, angle, and pose.}
Table~\ref{tab:mug-orientations} evaluates color selectivity across rotations and upright/inverted poses; * marks conditions only appear in the test set. The red mug yields mean SRs of 0–1.85\%, versus 64.85–99.10\% for other colors. Orange mugs show comparatively low SRs (64.85–82.85\%), possibly reflecting unintended activation due to their color similarity to the red trigger.

\noindent\textbf{Position robustness.} Figure~\ref{fig:trigger-positions} evaluates the red mug at eight insertion locations.
SR remains between 0\% and 9.72\%, indicating that the attack remains effective across the tested positions.

\noindent\textbf{Shape selectivity.} Figure~\ref{fig:distractors} holds color fixed in the same position while varying object shape.
Red distractors yield SRs of 81.25--100\%, whereas the red mug yields 0\%.
These results support selectivity for the trigger object beyond color alone.

\noindent\textbf{Object combinations.} Figure~\ref{fig:combinations} examines whether color and shape cues distributed across different objects can activate the attack.
A blue mug paired with a red capsule or box yields SRs of 95.37\% and 94.44\%, respectively, while the corresponding combinations containing the red mug both yield 2.78\%.
These results suggest that the selector responds to features of the trigger object, rather than merely to the co-occurrence of red color and mug shape across separate objects.

\begin{samepage}
\noindent\textbf{Trigger size.} Figure~\ref{fig:trigger-size}. Reducing the trigger size to \(0.5\times\) still yields an ASR of 81.03\% and an SG of 75.11\%.
Increasing the size to \(1.5\times\) slightly improves ASR from 90.22\% to 91.68\%, but reduces SG from 84.60\% to 83.22\%.
The \(1\times\) baseline therefore offers a more balanced trade-off between attack effectiveness and stealthiness.
\par
\end{samepage}

\raggedbottom
\section{Ablation}
\label{sec:ablation}

\subsection{Perturbation propagation}
\label{sec:propagation-ablation}

To trace how the selector–amplifier perturbation reaches the action head, we conduct activation restoration experiments on 500 trigger-present validation observations. Inputs follow the construction in Section 4.1.

\begin{samepage}
\noindent We evaluate each intervention independently, restoring selected activations to those computed by the original model on the same input. Table~\ref{tab:propagation_ablation} reports mean action MAE as defined in Eq.~(2).
\par
\end{samepage}

In Llama block 0, we separately restore visual keys or values immediately after projection and before attention computation. We then recompute attention and all subsequent layers. Restoring keys reduces action MAE from 0.21760 to 0.00255, whereas restoring values yields a deviation of 0.10175. This identifies altered visual keys, which determine how other tokens attend to visual content, as a critical component of the perturbation mechanism.

\begingroup
\widowpenalty=10000
\brokenpenalty=10000
\looseness=-1
At the output of the block 0, we retain the modified hidden states only at text positions and restore all other positions to their original states. This retains an action MAE of 0.21737. Conversely, restoring only the text states while retaining all other modified states reduces MAE to 0.00458. Retaining only the modified visual states yields an MAE of 0.00288. These results show that, after the first block, text representations carry much of the effect that subsequently reaches the action head, connecting the early visual perturbation to the final action deviation.
\par
\endgroup

\subsection{Bit budget analysis}
Figure~\ref{fig:bit-budget} evaluates different selector budgets with the amplifier budget fixed at 8 bits. With only 30 selector bit flips, trigger SR falls below 10\%, while clean SR remains near baseline and distractor SR exceeds 85\%. Increasing the budget further yields smaller gains in trigger suppression and generally improves distractor SR.

\newsavebox{\BitBudgetPanel}
\sbox{\BitBudgetPanel}{%
\begin{minipage}{\columnwidth}
\captionsetup{skip=5pt,hypcap=false}
\centering
\includegraphics[width=\linewidth]{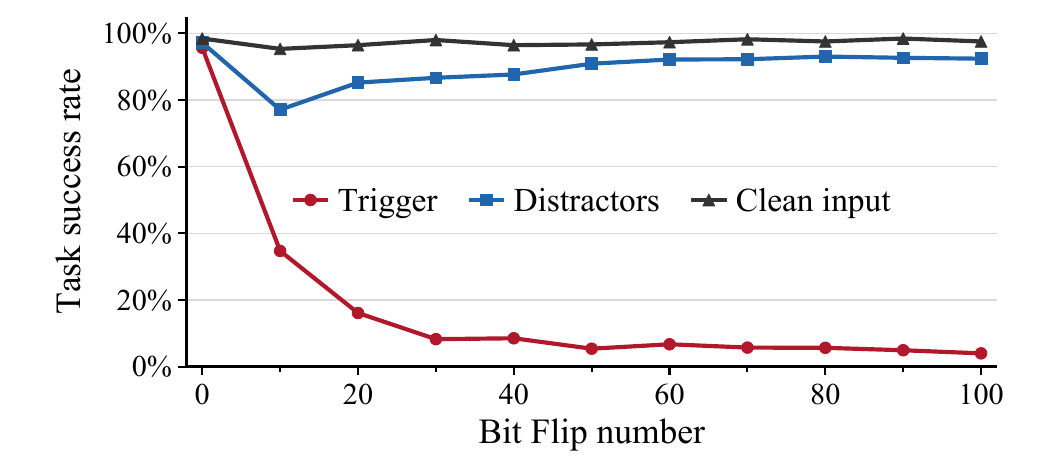}
\captionof{figure}{\textbf{Bit-Budget Analysis.}
Task success rates on clean, trigger and distractor scenes for OpenVLA-OFT on LIBERO-Spatial. The selector budget varies, while the amplifier budget is fixed at 8 bits ($\lambda=2$).}
\label{fig:bit-budget}
\end{minipage}}
\par\vspace{12.8pt}
\noindent\usebox{\BitBudgetPanel}\par
\vspace{4pt}

\makeatletter
\setlength{\@fptop}{0pt}
\makeatother

\setlength{\textfloatsep}{6pt plus 2pt minus 2pt}
\suppressfloats[t]
\begin{table}[t]
\centering
\caption{\textbf{Ablations of Perturbation Propagation.}}
\label{tab:propagation_ablation}
\begingroup
\definecolor{PropRestore}{RGB}{34,76,112}
\fontsize{9}{10.5}\selectfont
\setlength{\tabcolsep}{3pt}
\renewcommand{\arraystretch}{1.69}
\begin{minipage}{\columnwidth}
\textbf{(a) Visual Key/Value Restoration}\par\vspace{3pt}
\begin{tabular*}{\linewidth}{@{\extracolsep{\fill}}llrr@{}}
\toprule
\textbf{Visual K} & \textbf{Visual V} &
\shortstack[r]{\textbf{Action}\\\textbf{MAE}} &
\shortstack[r]{\textbf{Relative}\\\textbf{MAE (\%)}} \\
\midrule
Modified & Modified & 0.21760 & 100.0 \\
\midrule
\textcolor{PropRestore}{Original} & Modified & \textbf{0.00255} & \textbf{1.2} \\
Modified & \textcolor{PropRestore}{Original} & 0.10175 & 46.8 \\
\bottomrule
\end{tabular*}
\par\vspace{8pt}
\textbf{(b) Token Representation Restoration}\par\vspace{3pt}
\begin{tabular*}{\linewidth}{@{\extracolsep{\fill}}lllrr@{}}
\toprule
\shortstack[l]{\textbf{Text}\\\textbf{tokens}} &
\shortstack[l]{\textbf{Visual}\\\textbf{tokens}} &
\shortstack[l]{\textbf{Other}\\\textbf{tokens}} &
\shortstack[r]{\textbf{Action}\\\textbf{MAE}} &
\shortstack[r]{\textbf{Relative}\\\textbf{MAE (\%)}} \\
\midrule
Modified & Modified & Modified & 0.21760 & 100.0 \\
\midrule
Modified & \textcolor{PropRestore}{Original} & \textcolor{PropRestore}{Original} & \textbf{0.21737} & \textbf{99.9} \\
\textcolor{PropRestore}{Original} & Modified & Modified & 0.00458 & 2.1 \\
\textcolor{PropRestore}{Original} & Modified & \textcolor{PropRestore}{Original} & 0.00288 & 1.3 \\
\bottomrule
\end{tabular*}
\end{minipage}
\endgroup
\end{table}

\subsection{Effect of the suppression weight \texorpdfstring{\(\lambda\)}{lambda}}
\label{sec:suppression-weight}

\begingroup
\looseness=-1
\noindent We examine how the suppression weight \(\lambda\) in Eq.~\eqref{eq:selector-objective} affects the activation separation achieved by the selector on the same LIBERO-Spatial dataset.
We report the 5th percentile (P5) of peak activations after GELU on positive observations and the 95th percentile (P95) on negative observations, with a positive \(\mathrm{P5}-\mathrm{P95}\) gap indicating good activation separability.
\par
\endgroup

\begingroup
\looseness=-1
\noindent
Figure~\ref{fig:lambda-suppression} shows that the difference between positive P5 and negative P95 increases with the FC2 bit budget for all tested values of \(\lambda\).
With \(\lambda=1\), negative P95 remains relatively high, while \(\lambda=10\) strongly suppresses negative responses but also reduces positive P5.
At 60 FC2 bit flips, \(\lambda=3\) yields the largest gap, whereas \(\lambda=2\) yields the largest gaps at \mbox{80--120} bits, highlighting the trade-off between trigger activation and background suppression under different bit budgets.
\par
\endgroup

\par\vspace{2.55pt}
\noindent\begin{minipage}{\columnwidth}
\centering
\captionsetup{skip=5pt,hypcap=false}
\includegraphics[width=\linewidth]{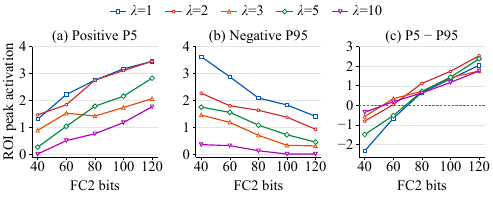}
\captionof{figure}{Effect of the suppression weight \(\lambda\).}
\label{fig:lambda-suppression}
\end{minipage}\par
\vspace{6pt}

\section{Related Work}
\label{sec:related-work}

Bit-flip attacks modify deployed model parameters without retraining~\cite{rakin2019bfa}, inducing degradation~\cite{hong2019brain,yao2020deephammer}, targeted errors~\cite{rakin2022tbfa,bai2021targeted}, or backdoor behavior~\cite{rakin2020tbt,chen2021proflip,li2025oneflip} in neural networks. Hardware studies~\cite{kim2014rowhammer,veen2016drammer,frigo2018grand}, including Rowhammer on GPU memory~\cite{lin2025gpuhammer,hu2026gddrhammer,wan2026geforge,lin2026gpubreach}, motivate this attack surface. Related work on LLMs and agents extends these concerns to generation~\cite{xu2025silentstriker,lin2026groundhog}, safety alignment~\cite{coalson2024prisonbreak}, and tool use~\cite{wang2026agents}.

VLA research improves robot control through large-scale training~\cite{brohan2023rt1,driess2023palme,zitkovich2023rt2,openx2024,ghosh2024octo,kim2025openvla}, diverse action representations~\cite{black2025pi0,liu2025rdt,li2024cogact,black2025pi05,qu2025spatialvla,pertsch2025fast}, and efficient adaptation~\cite{kim2025oft,shukor2025smolvla,bjorck2025groot,nvidia2026groot17}. Benchmarks support systematic evaluation~\cite{liu2023libero,li2025simpler}. Existing VLA backdoors rely on poisoned data or malicious fine-tuning to induce trigger-dependent behavior~\cite{zhou2025badvla,xu2026dropvla}. 

\begin{samepage}
\section{Conclusion}

FlipVLA implants object-triggered backdoors into VLA models without retraining through a selector--amplifier construction using limited parameter bit flips. Across two models and four LIBERO suites, it achieves 90.22\%--100\% ASR with at most a 1.5 percentage point decrease in clean-task success. These results show that fixed parameter corruption can induce conditional task failures, motivating evaluation beyond clean-input performance.
\par
\end{samepage}

\clearpage
\section*{LLM Usage Statement}

We used OpenAI Codex to assist with manuscript editing,
LaTeX typesetting, figure and table preparation,
recalculating ASR and SG from author-provided rollout
success rates, and organizing and checking references.
The authors reviewed and verified all AI-assisted content,
including numerical calculations and citations, and take
full responsibility for the accuracy, originality, and
integrity of the final manuscript.

{
    \small
    \urlstyle{same}
    \bibliographystyle{ieeenat_fullname}
    \bibliography{main,sec/references}
}

\ifincludesupplementary
\onecolumn
\setcounter{page}{1}
\renewcommand{\thepage}{S\arabic{page}}
\raggedbottom
{\centering\Large
 \textbf{\thetitle}\par
 \vspace{.5em}Supplementary Material\par
 \vspace{15.4pt}}

\appendix

\setcounter{figure}{0}
\setcounter{table}{0}
\renewcommand{\thefigure}{S\arabic{figure}}
\renewcommand{\thetable}{S\arabic{table}}
\renewcommand{\theHfigure}{supp.\arabic{figure}}
\renewcommand{\theHtable}{supp.\arabic{table}}

\setlength{\multicolsep}{0pt}
\begin{multicols}{2}
\raggedcolumns
\section{Additional experimental details}
\label{app:experimental-details}

\noindent\textbf{Attack settings.}
We use a red
mug as the trigger throughout the main experiments across both models
and all four suites. Additional trigger types are evaluated in
Section~\ref{app:trigger-replacement}. Attacks are constructed separately
for each model--suite pair; these experiments assess the applicability
of FlipVLA across the tested settings.

\noindent\textbf{Attack-construction data.}
For the main OpenVLA-OFT experiment on LIBERO-Spatial, we partition the
attack-construction data by source demonstration within each task. We use
ten demonstrations per task for optimization and five disjoint
demonstrations for validation, sampling observations at 0\%, 33\%, and
66\% of each trajectory. Across ten tasks, this yields 300 optimization
groups and 150 validation groups. Each group contains one clean
observation, 24 trigger-present variants, and 63 distractor-only variants.
All observations and variants derived from a demonstration remain in the
same partition. The optimization and validation sets also use separate
insertion-position pools. The validation set supports host-channel
selection.

\noindent\textbf{Optimization objectives.}
The selector promotes trigger activation while suppressing background
responses across clean, trigger-present, and distractor-only inputs.
With the selector frozen, amplifier optimization maximizes mean action
MAE on the positive set alone, without additional MAE objectives or
constraints on clean or distractor inputs. This division assigns trigger
discrimination and background suppression to the selector, while the
amplifier increases the action deviation induced by its responses.

\noindent\textbf{Paired rollout evaluation.}
We evaluate the original and bit-flipped policies on paired scenarios
under clean, trigger-present, and distractor-only conditions. Each policy
pair uses the same task instruction, initial state, and inserted-object
configuration.

\columnbreak
\noindent\begin{minipage}{\columnwidth}
  \captionsetup{type=table,position=top,hypcap=false}
  \centering
  \caption{\textbf{Component ablation for GR00T N1.7.}}
  \label{tab:supp-groot-components}
  \begingroup
  \footnotesize
  \setlength{\tabcolsep}{1.35pt}
  \setlength{\fboxsep}{1pt}
  \renewcommand{\arraystretch}{1.12}
  \begin{tabular*}{\columnwidth}{@{\extracolsep{\fill}}lrrrrr@{}}
    \toprule
    \textbf{Version} & \textbf{SR (+)} & \textbf{SR ($-$)} & \textbf{SR (c)} &
      \textbf{ASR}~$\uparrow$ & \textbf{SG}~$\uparrow$ \\
    \midrule
    \rlap{\color[gray]{0.94}\rule[-0.40em]{\columnwidth}{1.33em}}%
    Clean(Baseline)      & 95.00 & 97.00 & 96.50 & -- & -- \\
    Selector-only        & 95.00 & 95.25 & 97.00 & 0.00 & $-1.80$ \\
    Amplifier-only       & 94.75 & 96.75 & 97.50 & 0.27 & 0.01 \\
    Selector + Amplifier & \textcolor{red}{7.50} & 96.25 & 97.00 &
      \colorbox{red!6}{\textbf{92.58}} & \colorbox{red!6}{\textbf{91.33}} \\
    \bottomrule
  \end{tabular*}\par
  \endgroup
  \smallskip
  \parbox{\columnwidth}{\footnotesize\raggedright\textit{Note.} All metrics are
  percentages. For each variant, the 1,000 rollouts comprise 400 trigger-present~$(+)$,
  400 distractor-only~$(-)$, and 200 clean~(c) episodes.}
\end{minipage}\par
\medskip
\noindent The main Spatial evaluation covers 151 configurations
across ten tasks, totaling 18,500 episodes per policy: 4,320
trigger-present, 13,680 distractor-only, and 500 clean episodes. Each
inserted-object condition is evaluated on six initial states per task,
with one seen and one unseen insertion location per initial state. Clean
evaluation uses 50 initial states per task.

\section{Component ablation}
\label{app:component-ablation}

We use the same GR00T N1.7 checkpoint as in the main LIBERO-Spatial
evaluation. Each single-component variant retains only its corresponding
subset of the 68 modified bits (Table~\ref{tab:supp-groot-components}).
The original model and the selector-only variant each succeed in 380 of
400 trigger-present rollouts, compared with 379 for the amplifier-only
variant. With both components present, trigger SR falls from 95.00\%
to 7.50\% (30/400), while clean and distractor SRs remain within
0.75 percentage points of their respective baselines. Neither component
alone reproduces this stealthy attack behavior, suggesting tight
functional coupling between the selector and amplifier in this setting.

\section{Trigger replacement}
\label{app:trigger-replacement}

We compare the default red mug with a blue tetrahedron and a green cube
as triggers for GR00T N1.7 on LIBERO-Spatial
(Fig.~\ref{fig:supp-trigger-objects}).
\end{multicols}

\medskip
\noindent\begin{minipage}{\textwidth}
  \captionsetup{type=figure,hypcap=false}
  \centering
  \includegraphics[
    width=\textwidth,
    trim=22bp 201bp 22bp 200bp,
    clip
  ]{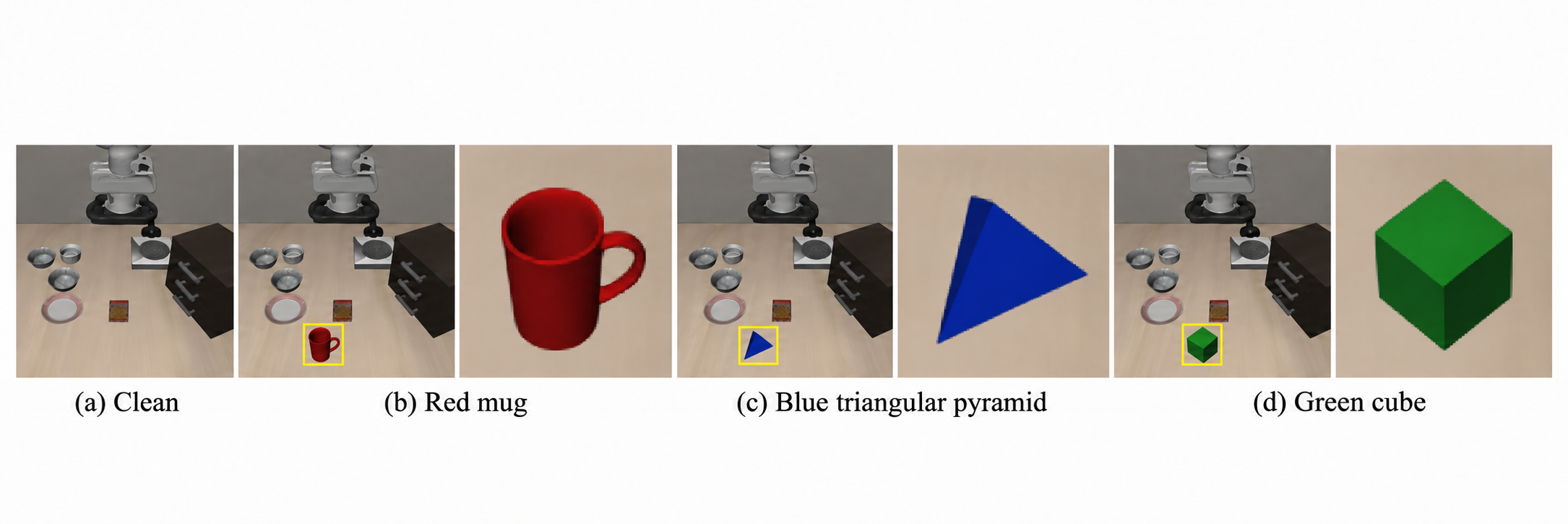}
  \par\vspace{2pt}
  {\small\noindent
    \makebox[0.142\textwidth][c]{(a) Clean}%
    \makebox[0.286\textwidth][c]{(b) Red mug}%
    \makebox[0.286\textwidth][c]{(c) Blue tetrahedron}%
    \makebox[0.286\textwidth][c]{(d) Green cube}\par
  }
  \caption{\textbf{Trigger objects on LIBERO-Spatial.}
    The leftmost panel shows the clean scene.
    The remaining panels show the red mug, blue tetrahedron, and green cube
    from left to right.
    Each object is shown in the scene (yellow box) and in the adjacent close-up.}
  \label{fig:supp-trigger-objects}
\end{minipage}\par

\twocolumn
\noindent
Each trigger is evaluated with its own modified model.
All three experiments use the same hyperparameters, with 300 observation
 groups for optimization and 150 for validation per trigger.
Trigger SR falls to 0.50\% for the blue tetrahedron and 0.25\% for the green
cube, compared with 7.50\% for the red mug
(Table~\ref{tab:supp-groot-trigger-replacement}).
In the paired trigger-present rollouts, all outcome changes are from success
to failure (389 for the tetrahedron and 386 for the cube).
For both replacement triggers, aggregate clean SR matches its original value,
and distractor SR stays at or above its original value.

\begin{table}[!htbp]
  \centering
  \caption{\textbf{Effect of trigger choice on GR00T N1.7.}
    Results for the red mug are from the main table.}
  \label{tab:supp-groot-trigger-replacement}
  \begingroup
  \footnotesize
  \setlength{\tabcolsep}{1.35pt}
  \setlength{\fboxsep}{1pt}
  \renewcommand{\arraystretch}{1.12}
  \newcommand{\grootOriginalShade}{%
    \rlap{\color[gray]{0.94}%
      \rule[-0.40em]{\columnwidth}{1.33em}}}
  \newcommand{\grootMetric}[1]{\colorbox{red!6}{\textbf{#1}}}
  \begin{tabular*}{\columnwidth}{@{\extracolsep{\fill}}llrrrrr@{}}
    \toprule
    \textbf{Trigger} & \textbf{Version} &
      \textbf{SR (+)} & \textbf{SR ($-$)} & \textbf{SR (c)} &
      \textbf{ASR}~$\uparrow$ & \textbf{SG}~$\uparrow$ \\
    \midrule
    \grootOriginalShade Red mug & Original &
      95.00 & 97.00 & 96.50 & -- & -- \\
      & Bit-flipped & \textcolor{red}{7.50} & 96.25 & 97.00 &
        \grootMetric{92.58} & \grootMetric{91.33} \\
    \midrule
    \grootOriginalShade Blue tetrahedron & Original &
      97.75 & 96.00 & 98.50 & -- & -- \\
      & Bit-flipped & \textcolor{red}{0.50} & 97.25 & 98.50 &
        \grootMetric{99.49} & \grootMetric{100.79} \\
    \midrule
    \grootOriginalShade Green cube & Original &
      96.75 & 96.00 & 97.00 & -- & -- \\
      & Bit-flipped & \textcolor{red}{0.25} & 96.00 & 97.00 &
        \grootMetric{99.74} & \grootMetric{99.74} \\
    \bottomrule
  \end{tabular*}\par
  \endgroup
  \smallskip
  \parbox{\columnwidth}{\footnotesize\raggedright
    \textit{Note.} All metrics are percentages.
    Each trigger accounts for 2,000 rollouts, with 400 trigger-present~$(+)$,
    400 distractor-only~$(-)$, and 200 clean~$(c)$ episodes per model version.
    ASR and SG use the corresponding original model as the baseline.
    For the blue tetrahedron, SG exceeds 100\% because distractor SR
    increases from 96.00\% to 97.25\%.}
\end{table}

\fi

\end{document}